\documentclass[10pt,twocolumn,letterpaper]{article}

\usepackage{iccv}              % To produce the CAMERA-READY version
\definecolor{iccvblue}{rgb}{0.21,0.49,0.74}
\usepackage[pagebackref,breaklinks,colorlinks,allcolors=iccvblue]{hyperref}
\usepackage{hyperref}
\usepackage{url}
\usepackage{multirow,booktabs,graphicx,siunitx,wrapfig,subcaption}
\usepackage{colortbl}
\usepackage{xcolor}
\usepackage{multicol}
\usepackage{balance}
\def\paperID{3685} % *** Enter the Paper ID here
\def\confName{ICCV}
\def\confYear{2025}

\title{IGL-Nav: Incremental 3D Gaussian Localization for Image-goal Navigation}

\author{Wenxuan Guo\textsuperscript{1$*$} \quad
Xiuwei Xu\textsuperscript{1$*$} \quad
Hang Yin\textsuperscript{1} \quad
Ziwei Wang\textsuperscript{2} \\
Jianjiang Feng\textsuperscript{1$\dagger$} \quad
Jie Zhou\textsuperscript{1} \quad
Jiwen Lu\textsuperscript{1} \\
	\textsuperscript{1}Tsinghua University \qquad \textsuperscript{2}Nanyang Technological University\\
	{\tt\small\{gwx22, xxw21, yinh23\}@mails.tsinghua.edu.cn \quad ziwei.wang@ntu.edu.sg} \\
	{\tt\small \{jfeng, jzhou, lujiwen\}@tsinghua.edu.cn}}

\begin{document}
\maketitle
{
        \renewcommand{\thefootnote}{*}
	\footnotetext[5]{Equal contribution. $^{\dagger}$ Corresponding author.}
}

\begin{abstract}
    Visual navigation with an image as goal is a fundamental and challenging problem.
Conventional methods either rely on end-to-end RL learning or modular-based policy with topological graph or BEV map as memory, which cannot fully model the geometric relationship between the explored 3D environment and the goal image. 
In order to efficiently and accurately localize the goal image in 3D space, we build our navigation system upon the renderable 3D gaussian (3DGS) representation. However, due to the computational intensity of 3DGS optimization and the large search space of 6-DoF camera pose, directly leveraging 3DGS for image localization during agent exploration process is prohibitively inefficient.
To this end, we propose IGL-Nav, an Incremental 3D Gaussian Localization framework for efficient and 3D-aware image-goal navigation.
Specifically, we incrementally update the scene representation as new images arrive with feed-forward monocular prediction. Then we coarsely localize the goal by leveraging the geometric information for discrete space matching, which can be equivalent to efficient 3D convolution. When the agent is close to the goal, we finally solve the fine target pose with optimization via differentiable rendering.
The proposed IGL-Nav outperforms existing state-of-the-art methods by a large margin across diverse experimental configurations. It can also handle the more challenging free-view image-goal setting and be deployed on real-world robotic platform using a cellphone to capture goal image at arbitrary pose. Project page: \href{https://gwxuan.github.io/IGL-Nav/}{https://gwxuan.github.io/IGL-Nav/}. 
\end{abstract}

\section{Introduction}   
    \begin{figure}
% \vspace{-.3cm}
  \centering
  \includegraphics[width=0.47\textwidth]{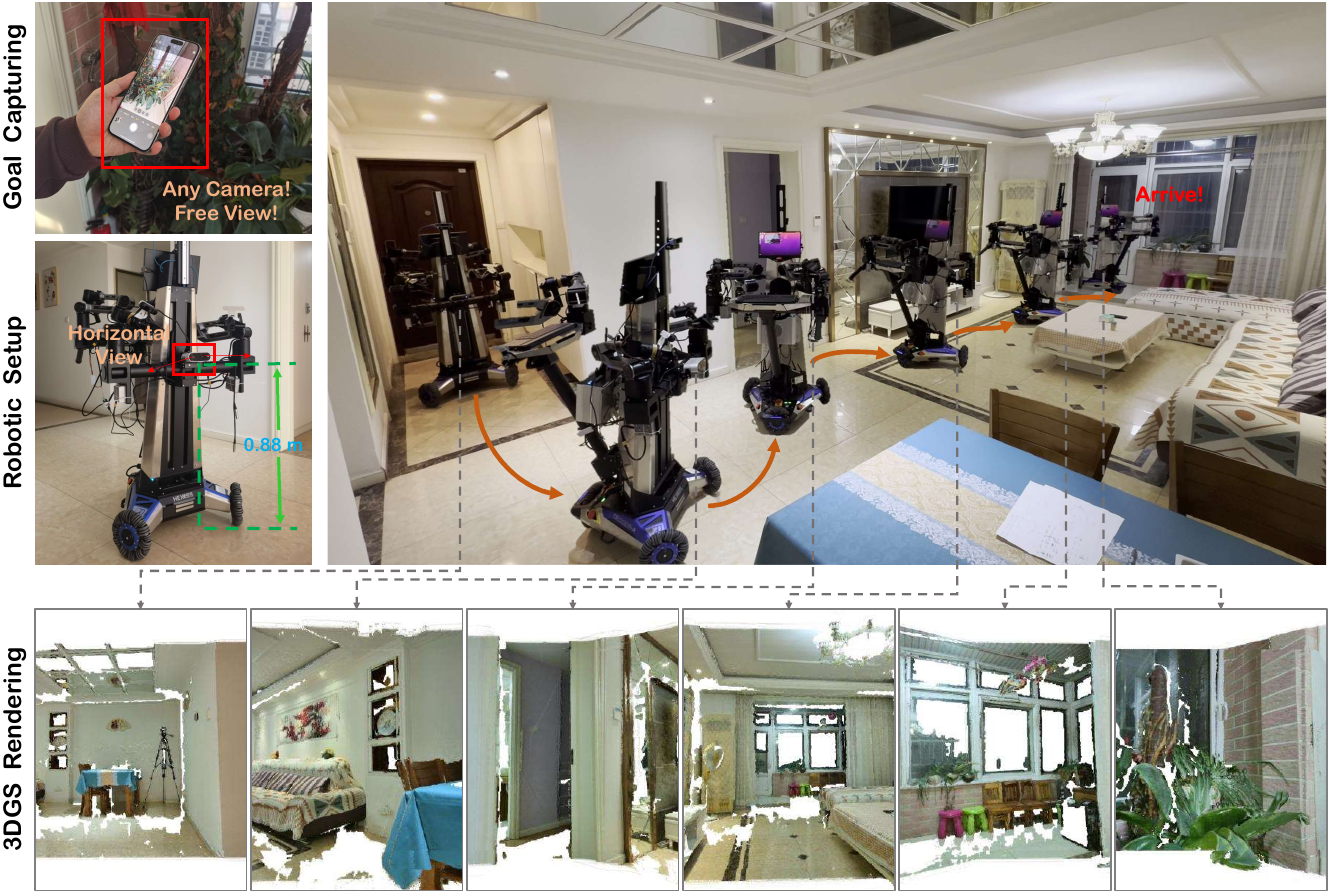}
  \vspace{-.15cm}
  \caption{IGL-Nav effectively guides the agent to reach free-view image goal via incremental 3D gaussian localization.}
  \label{fig:teaser}
  \vspace{-.3cm}
\end{figure}

Image-goal navigation, which requires an agent initialized in unknown environment to navigate to the location and orientation specified by an image~\cite{zhu2017target}, is a fundamental problem in a wide range of robotic tasks.
This task requires the agent to precisely understand spatial information, as well as to reason how to explore the scene with past observations, which is hard to learn with 
end-to-end RL~\cite{yadav2023ovrl} due to low sample efficiency and catastrophic forgetting.

% modular-based method-->explicit memory cache22
Recent advances in visual navigation have witnessed significant progress in modular-based approaches~\cite{chaplot2020object,chaplot2020neural,yin2024sg,yin2025unigoal,hahn2021no,ramakrishnan2022poni,kwon2023renderable}, which establish an explicit memory to cache observed environmental information and derive navigation policies based on the memory representations.
While these approaches demonstrate enhanced capabilities for long-horizon reasoning and temporal dependency modeling in object-goal navigation tasks, their extension to image-goal navigation remains challenging. Unlike object-goal scenarios that primarily rely on high-level semantic understanding, image-goal navigation necessitates the preservation and processing of low-level visual features, including fine-grained texture patterns and color distributions. Consequently, conventional representation paradigms of memory such as topological graphs prove insufficient for effectively encoding the requisite environmental information in image-goal settings.
To address these limitations, RNR-Map~\cite{kwon2023renderable} introduces a renderable neural radiance map representation. Drawing inspiration from NeRF~\cite{mildenhall2020nerf}, this representation enables photorealistic image rendering from arbitrary camera viewpoints. The renderable nature ensures the preservation of crucial low-level visual features, which has demonstrated superior performance in image-goal navigation.
However, since NeRF is an implicit field with high computational cost, RNR-Map has to maintain the renderable representation in a 2D BEV map for efficient and explicit memory management. This 2D projection inherently loses critical 3D structural information, forcing RNR-Map to impose strict constraints on goal image acquisition, specifically requiring horizontal camera angles to ensure alignment with its BEV map. This significantly reduces its applicability in real-world scenarios.
Therefore, an efficient 3D-aware memory representation is still desirable for image-goal navigation.

In this paper, we propose to leverage 3D Gaussian Splatting (3DGS)~\cite{kerbl20233d} as the scene representation for image-goal navigation.  
The 3DGS representation demonstrates exceptional suitability for the task: (1) as an explicit representation, 3DGS can be easily initialized with the observed RGB-D image and be incrementally accumulated in 3D space; (2) it supports efficient differentiable rendering, which can be used to localize the camera pose of goal image with iterative optimization.
Despite these compelling properties, adapting 3DGS representations for image-goal navigation presents significant challenges. While 3DGS achieves rendering speeds orders of magnitude faster than NeRF, their optimization process remains computationally prohibitive for real-time online inference required in navigation tasks. Furthermore, goal image localization within scene-level 3DGS maps becomes intractable due to the exponential search space complexity inherent in 6-DoF camera pose estimation.
To this end, we propose IGL-Nav, an \textbf{I}ncremental 3D \textbf{G}aussian \textbf{L}ocalization framework that (1) progressively constructs 3DGS through feed-forward prediction, eliminating offline optimization; and (2) enables efficient hierarchical goal search by harnessing both geometric and photometric attributes of 3DGS through our novel coarse-to-fine localization strategy.
Extensive experiments on various datasets in Habitat simulator show our IGL-Nav significantly outperforms previous state-of-the-art image-goal navigation methods. 
Moreover, benefit from our explicit 3D representation, IGL-Nav is also able to handle the more practical free-view image-goal setting, where there is no assumption on both camera intrinsics and extrinsics of the goal image. We further deploy our method on real-world robot, where a casually taken photo from a cellphone can be used as goal to guide the agent navigating to specified location in complicated and large-scale environments.

\section{Related Work}   
    % \cite{yin2024sg}
\textbf{3D Gaussian Splatting}.
3DGS~\cite{kerbl20233d} has emerged as a powerful technique for 3D scene representation. It represents a scene as a dense set of points with gaussian embedding and leverages efficient rasterization techniques for high-fidelity, real-time rendering. Recently, feed-forward 3DGS models~\cite{charatan2024pixelsplat,chen2024lara,chen2024mvsplat} have been proposed, primarily to address the issue of sparse-view scene reconstruction. Unlike traditional methods that iteratively optimize 3DGS parameters, feed-forward 3DGS predicts the gaussian distribution through a network, significantly improving modeling efficiency. 
In embodied AI, 3DGS has been applied to manipulation tasks, with dynamic 3DGS frameworks and gaussian world models used to model and predict robotic actions~\cite{lu2024manigaussian}. And systems like Gaussian-Grasper~\cite{zheng2024gaussiangrasper} leverage RGB-D inputs for language-guided grasping. 3DGS also helps bridge the gap between simulated and real-world environments for generalizing learned behaviors. Techniques such as Robo-GS~\cite{lou2024robo} and SplatSim~\cite{qureshi2024splatsim} improve Sim-to-Real transfer by leveraging efficient representation of 3DGS.
The incremental 3DGS scene representation used in our IGL-Nav also follows the paradigm of feed-forward 3DGS, making it suitable for scene modeling based on online inputs in navigation tasks.

\noindent\textbf{Image-goal Navigation}.
Image-goal navigation involves the agent navigating to the location where the goal image is captured~\cite{zhu2017target}, requiring precise alignment in both position and orientation. To address this challenge, researchers have employed various strategies. Some focus on optimizing reinforcement learning (RL) policies~\cite{yadav2023ovrl,al2022zero,yadav2023offline} that directly map observations to actions. Others concentrate on constructing detailed maps~\cite{chaplot2020neural,kim2023topological,kwon2023renderable,savinov2018semi}, or on developing carefully crafted matching algorithms~\cite{wasserman2023last}. 
However, image-goal navigation requires that the query image be captured by the agent's camera, which limits the camera's intrinsic parameters, height, and the fact that it can only rotate around the Z-axis. Considering these constraints in practical applications, we propose free-view image-goal navigation, where the target image can be captured by any camera at free 3D position and orientation.
In addition, instance image-goal navigation~\cite{krantz2022instance} is a similar task, where the target image focuses on specific categories of objects within the scene.
In instance image-goal task, GaussNav~\cite{lei2025gaussnav} also uses a 3DGS-based scene representation. However, GaussNav requires first completing the exploration of the entire building to optimize the 3DGS representation, and then render images at multi poses for comparison with the target image. This approach limits efficiency in practical applications. In contrast, our IGL-Nav simultaneously performs exploration, incremental modeling, and target localization, and it incorporates a coarse-to-fine localization strategy, making full use of the 3DGS representation.

\section{Approach}   
    In this section, we first describe our task definition. Next, we explain several core modules of IGL-Nav, including incremental scene representation with 3DGS, and coarse-to-fine target localization. Finally, we detail the overall navigation pipeline.

\begin{figure*}
	\centering
	\includegraphics[width=\linewidth]{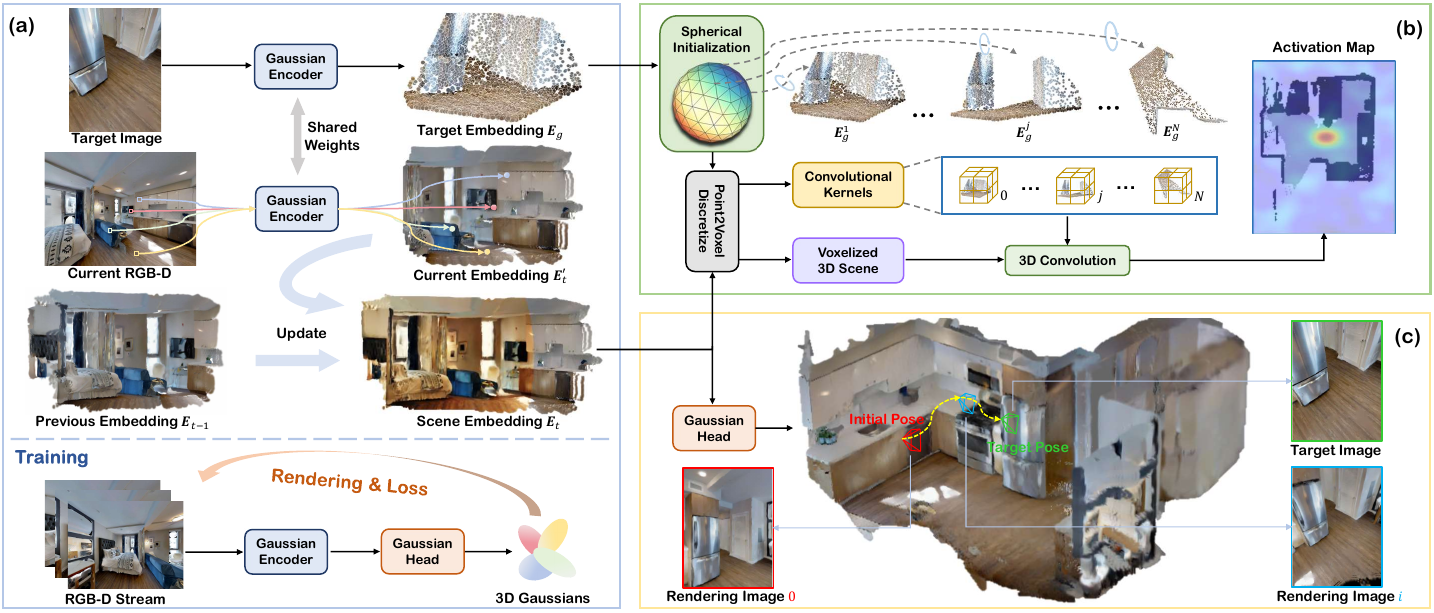}
    \vspace{-.6cm}
    \caption{Illustration of IGL-Nav. (a) We maintain an incremental 3DGS scene representation with feed-forward prediction. (b) The coarse target localization is modeled as a 5-dimension matching problem, which is efficiently implemented by leveraging the target embedding as 3D convolutional kernel. (c) Fine target localization via differentiable 3DGS rendering and matching-constrained optimization.}
	\label{fig:method1}
	\vspace{-.25cm}
\end{figure*}

\subsection{Problem Statement}\label{sec:3.1}
We study the problem of free-view image-goal navigation, which is a more challenging and practical setting.
In this task, a mobile agent is instructed with navigating to a specified location depicted by an image $\boldsymbol{I}_g$, taken by camera A with pose $\boldsymbol{\rm T}_g$. 
The agent is equipped with camera B. It receives posed RGB-D video stream $\{\boldsymbol{I}_t,\boldsymbol{D}_t,\boldsymbol{\rm T}_t\}_{t=1}^T$ and is required to execute an action $a\in \mathcal{A}$ at each time it receiving a new RGB-D observation. 
$\boldsymbol{I}_t,\boldsymbol{D}_t,\boldsymbol{\rm T}_t$ refer to RGB image, depth image and camera pose at time instant $t$.
$\mathcal{A}$ is the set of actions, which consists of \texttt{move\_forward}, \texttt{turn\_left}, \texttt{turn\_right} and \texttt{stop}. The task is considered successfully completed if the agent terminates within a horizontal neighborhood of the target pose, satisfying $||\mathcal{P}(\boldsymbol{\rm T}_{final})-\mathcal{P}(\boldsymbol{\rm T}_g)||_2<\epsilon$ within a maximum of $T$ navigation steps. Here $\mathcal{P}$ refers to 3D-to-BEV projection.

\textbf{Comparison with Relevant Tasks}.
In the free-view image-goal navigation, there is no assumption on the correlation between camera A and B. For example, in real application scenarios, A can be a cellphone, and B is a RGB-D camera with totally different camera intrinsics.
Previous image-goal setting~\cite{hahn2021no,kwon2023renderable} can be regarded as a special case of our task where $A\equiv B$ and $\boldsymbol{\rm T}_g$ is restricted to lie within camera B's achievable pose space.
Instance-image-goal navigation~\cite{krantz2023navigating} also aims to decouple camera A and B. However, this setting requires that there must be an instance located at image center, and only 6 categories of instances are supported. These limitations fundamentally constrain the system's operational flexibility and real-world deployment potential.
In this paper, we conduct experiments on both conventional and free-view image-goal settings for a comprehensive evaluation of different approaches.

\subsection{Incremental Scene Representation}\label{sec:3.2}
We adopt 3DGS as our scene representation due to its explicit nature and efficient rendering capability.
However, the original 3DGS are obtained through offline optimization on image set and thus hard to be applied in real-time tasks.
Recent feed-forward methods~\cite{charatan2024pixelsplat,chen2024lara,chen2024mvsplat} abandon optimization and directly predict pixel-aligned 3DGS parameters, but they still rely on multi-view images to reconstruct geometric information of the scene.
In visual navigation, the agent needs to incrementally build scene representation along with its exploration, so the 3DGS should be generated in real-time and update as new images arrive.
To accommodate streaming video input while effectively leveraging camera pose and depth priors, we present the first feed-forward 3DGS reconstruction model for monocular RGB-D sequences, which supports real-time 3DGS reconstruction and incremental accumulation.

\textbf{Gaussian Parameters Prediction}.
At time step $t$, the agent receives new RGB-D observations $\boldsymbol{I}_t \in \mathbb{R}^{H \times W \times 3}$ and $\boldsymbol{D}_t \in \mathbb{R}^{H \times W \times 1}$. Our incremental reconstruction model is essentially a mapping $f_{\boldsymbol{\theta}}$ from observations to 3DGS parameters, including position $\boldsymbol{\mu}_k$, opacity $\alpha_k$, covariance $\boldsymbol{\Sigma}_k$ and spherical harmonics $\boldsymbol{c}_k$:
\begin{equation}
    f_{\boldsymbol{\theta}}:\left(\boldsymbol{I}_t, \boldsymbol{D}_t\right) \mapsto\left\{\left(\boldsymbol{\mu}_k, \alpha_k, \boldsymbol{\Sigma}_k, \boldsymbol{c}_k\right)\right\}_{k=1}^{H \times W}
\end{equation}
The 3DGS parameters are predicted in a pixel-aligned manner, thus an observation input of size $H \times W$ corresponds to an output of $H \times W$ gaussians.

\begin{figure}
% \vspace{-.3cm}
  \centering
  \includegraphics[width=0.47\textwidth]{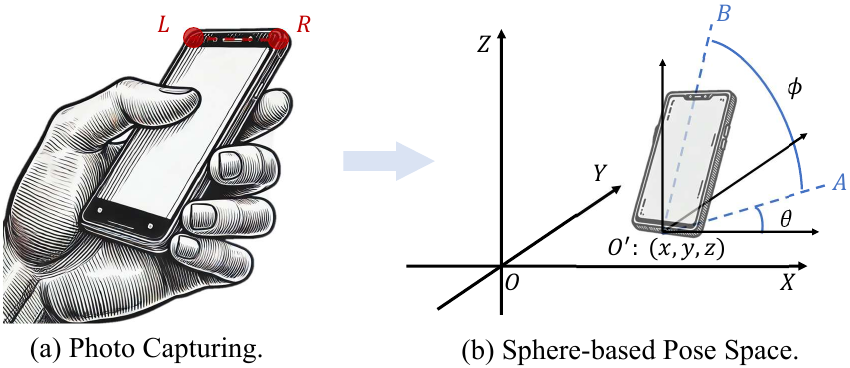}
  \vspace{-.2cm}
  \caption{Modeling of the camera pose space. (a) Line $LR$ is almost always parallel to the ground. (b) Line $AO'$ is parallel to Plane $XOY$. Plane $AO'B$ is perpendicular to Plane $XOY$.}
  \label{fig:method2}
  \vspace{-.3cm}
\end{figure}

The feed-forward model $f_{\boldsymbol{\theta}}$ is shown in Figure \ref{fig:method1}. We first concatenate the normalized RGB and depth images, and then extract dense monocular scene embedding $\boldsymbol{E}'_t$ with a UNet-based encoder $\mathcal{E}$. Then 3DGS parameters are regressed through a gaussian head $\mathcal{H}$, composed of a few CNN and linear layers. This process can be expressed as:
\begin{equation}
\Delta \boldsymbol{C}_{2D}, \Delta \boldsymbol{D}, \alpha, \boldsymbol{\Sigma},\boldsymbol{c} = \mathcal{H}(\boldsymbol{E}'), \quad \boldsymbol{E}' =  \mathcal{E}(\boldsymbol{I}, \boldsymbol{D})
\end{equation}
where $\Delta \boldsymbol{C}_{2D}$ and $\Delta \boldsymbol{D}$ are residuals of image coordinates and depth. We omit subscript $t$ for simplicity. Using the camera intrinsic matrix $\boldsymbol{\rm M}$, pose $\boldsymbol{\rm T}_t$ and inverse projection $\rm{Proj}^{-1}$, we can compute the 3DGS positions as:
\begin{equation}
\boldsymbol{\mu} = \rm{Proj}^{-1}(\boldsymbol{C}_{2D} + \Delta \boldsymbol{C}_{2D}, \boldsymbol{D} + \Delta \boldsymbol{D}\ |\ \boldsymbol{\rm M},\boldsymbol{\rm T}_t)
\end{equation}
We also lift $\boldsymbol{E}'$ from 2D to the corresponding 3D positions.
Finally, the 3DGS scene representation $\boldsymbol{G}$ and the corresponding 3D embedding $\boldsymbol{E}$ can be updated as:
$\boldsymbol{G}_t=\boldsymbol{G}_{t-1}\cup (\boldsymbol{\mu}_t,\alpha_t,\boldsymbol{\Sigma}_t,\boldsymbol{c}_t)$ and $\boldsymbol{E}_t=\boldsymbol{E}_{t-1}\cup \boldsymbol{E}'_t$. When the number of 3DGS in the scene is large, we prune $\boldsymbol{G}_t$ and $\boldsymbol{E}_t$ based on opacity and 3DGS density to reduce memory footprint.
Additionally, we can use $\mathcal{E}$ to extract the 3D embedding $\boldsymbol{E}_g$ of the target image $\boldsymbol{I}_g$. If depth and camera intrinsics are unavailable for $\boldsymbol{I}_g$, we simply use a monocular depth estimator~\cite{piccinelli2024unidepth} to predict them.

\textbf{Training and Loss}.
Our feed-forward model can be trained using passive offline RGB-D video streams. We randomly sample training episodes from navigation training set. In each episode, $K$ frames are randomly selected to predict 3DGS parameters, and images from other viewpoints are rendered for loss computation. The training loss is a linear combination of L-2 and LPIPS~\cite{zhang2018unreasonable} losses.

\subsection{Coarse-to-fine Localization}\label{sec:3.3}
Since the target image is captured by an arbitrary camera at any pose (6-DoF), the search space of the target is extremely large. To perform efficient and accurate visual navigation, we design a coarse-to-fine target localization strategy. 
Coarse localization leverages the incremental scene embedding $\boldsymbol{E}_t$ to predict the approximate target location in real-time during exploration. 
Once the agent is close to the target, fine localization is employed to accurately determine the accurate target position and guide the agent to reach it.

\subsubsection{Coarse Target Localization}
Although camera A can capture the target image at an arbitrary pose, we observe that the top frame of the camera is almost always parallel to the ground when taking a photo, as shown in Figure \ref{fig:method2}.
Therefore, we can represent the actual camera rotation with $(\theta,\phi)$, which denotes a rotation around the X-axis by $\theta$ degrees, followed by a rotation towards the Z-axis by $\phi$ degrees. 
Based on this observation, we define a sphere-based space $\mathcal{S}:\{(x,y,z,\theta,\phi)\}$ to represent camera pose. Here $(x,y,z)$ represents the position of camera A and $(\theta,\phi)$ refers to A's rotation.
We can thus represent the target pose $\boldsymbol{\rm T}_g$ as $(x_g,y_g,z_g,\theta_g,\phi_g)$. 
The 3D embedding of the target $\boldsymbol{E}_g$ is initialized at the origin of the sphere-based space $\mathcal{S}$ and should be aligned with the scene embedding $\boldsymbol{E}_t$ under translation $(x_g, y_g, z_g)$ and rotation $(\theta_g, \phi_g)$, which are unknown.

To efficiently search the target camera pose in the five-dimensional space, we discretize $\mathcal{S}$ to reduce the search space. For $(x,y,z)$, the 3D space is voxelized into grids $\{(x_i,y_i,z_i)\ |\ x_i=\lfloor \frac{x}{v}\rfloor,y_i=\lfloor \frac{y}{v}\rfloor,z_i=\lfloor \frac{z}{v}\rfloor\}$, where $v$ is voxel size. 
For $(\theta,\phi)$, we discretize the spherical surface into $N$ vertices of a hierarchical mesh via $\gamma$-level subdivision of a regular icosahedron, as shown in Figure \ref{fig:method1}.
In this way, we can rotate $\boldsymbol{E}_g$ according to the discretized sphere to obtain $N$ 3D embeddings $\{\boldsymbol{E}_{g}^1,...,\boldsymbol{E}_{g}^N\}$. By translating these embeddings to the discretized voxel grids and computing the extent of alignment between the translated embedding and $\boldsymbol{E}_t$, the coarse target pose can be determined by:
\begin{equation}\label{eq4}
    \mathop{\rm maximize}_{i,k}\ \ \mathcal{A}(\boldsymbol{E}_t,\mathcal{T}(\boldsymbol{E}_g^k,(x_i,y_i,z_i)))
\end{equation}
where $\mathcal{A}$ computes the extent of alignment between two sets of 3D features. $\mathcal{T}$ stands for translation operation. $i$ and $k$ are used to query the corresponding $(x,y,z,\theta,\phi)$.

However, the above operation is still hard to achieve real-time inference. We need to traverse all voxel grids and compare the translated 3D embedding with $\boldsymbol{E}_t$. Assume there are $V$ grids at all, then $V\times N$ times comparisons should be performed. Moreover, during each comparison, we should compute the geometric similarity between two 3D pointclouds as well as their feature similarity, which is especially time-consuming and hard to be accelerated on GPUs.
To solve this problem, we propose to further discretize the 3D embeddings $\boldsymbol{E}_t$ and $\boldsymbol{E}_g$. For $\boldsymbol{E}_t$, we can simply quantize the pointclouds into voxels, where voxel features are obtained by taking average of the pointcloud features inside each voxel. For $\{\boldsymbol{E}_{g}^1,...,\boldsymbol{E}_{g}^N\}$, we uniformly quantize them into $L\times L\times L$ voxels. Note that although $\{\boldsymbol{E}_{g}^1,...,\boldsymbol{E}_{g}^N\}$ are different pointclouds, they will share the same shape after voxelization, which forms a 3D convolutional kernel $\boldsymbol{K}\in\mathbb{R}^{L\times L\times L\times C_{in}\times C_{out}}$. Here $C_{in}$ refers to the output channel of $\mathcal{E}$, $C_{out}$ equals to the number of kernels $N$.
Therefore, Eq (\ref{eq4}) can be rewritten as:
\begin{equation}
    \mathop{\rm argmax}_{x,y,z,k}\ \ \mathcal{C}(f_1(\mathcal{V}(\boldsymbol{E}_t)),f_2(\boldsymbol{K}))[x][y][z][k]
\end{equation}
where $\mathcal{C}$ means 3D convolution operation, $\mathcal{V}$ quantizes scene embedding $\boldsymbol{E}_t$ into $X\times Y\times Z$ voxels. We use two MLP $f_1$ / $f_2$ with input channel $C_{in}$ and output channel $C'$ to project scene embedding and convolutional kernel to a learnable feature space before convolution, which further aligns the embedding space of $\boldsymbol{E}_t$ and $\boldsymbol{E}_g$.
The activation map after 3D convolution is of shape $X\times Y\times Z\times N$, from which we query index of the maximum value and thus obtain a coarse localization of the target pose.
To further improve computational efficiency, we use pillar-based voxelization~\cite{yin2021center,lang2019pointpillars}.

\textbf{Training and Loss}.
Similar to the scene representation training, we train the coarse localization module using offline passive video streams. In each training segment, we randomly select a position and capture target images with arbitrary intrinsic parameters and orientations. 
We use focal loss~\cite{lin2017focal} to supervise the activation map after 3D convolution. Additionally, we apply cross-entropy loss to supervise the outputs nearby target pose in the activation map.

\begin{figure}
% \vspace{-.3cm}
  \centering
  \includegraphics[width=0.47\textwidth]{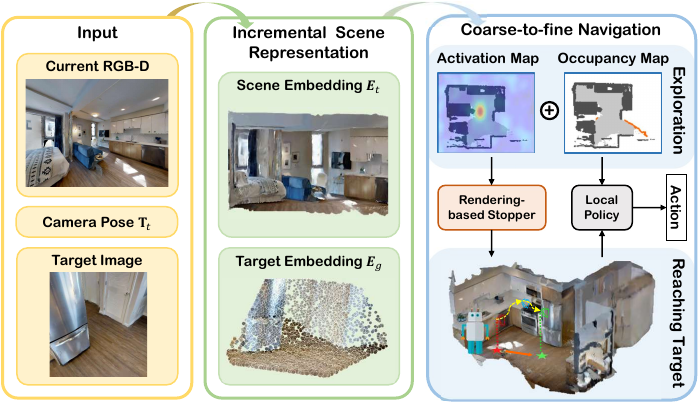}
  \vspace{-.23cm}
  \caption{Navigation pipeline of IGL-Nav.}
  \label{fig:method3}
  \vspace{-.4cm}
\end{figure}

\subsubsection{Fine Target Localization}
Our fine localization method aims to accurately determine the target's 6-DoF pose once the agent is close to the target region. It leverages the differentiable rendering ability of 3DGS to reach target pose via iterative optimization.

\textbf{Rendering-based Stopper}.
First, we use a rendering-based stopper to determine if the agent is close to the target. Since the intrinsics of camera A and B may differ significantly, directly comparing the current observation with the target image with feature matching is difficult. 
Thanks to the real-time rendering capability of 3DGS $\boldsymbol{G}_t$, we can render an image at camera B's current viewpoint with the same intrinsic parameters as camera A. We use a local feature matching method, LoFTR~\cite{sun2021loftr}, to predict matching pairs $(\boldsymbol{x}_g, \boldsymbol{x}_t)$ between the target image and the rendered image.
Here $(\boldsymbol{x}_g, \boldsymbol{x}_t)$ is the coordinate set of matched pixels.
If the number of matching pairs exceeds a threshold $\tau$, it is considered that $\boldsymbol{I}_g$ appears in agent's field of view.

\textbf{Matching-constrained Optimization}.
Via differentiable rendering, we can optimize the current camera pose with photometric loss between the rendered image and $\boldsymbol{I}_g$, and between rendered depth and $\boldsymbol{D}_g$ ($\boldsymbol{D}_g / \boldsymbol{\rm M}_g$ are the depth / intrinsics of $\boldsymbol{I}_g$, which are estimated by \cite{piccinelli2024unidepth} if not available), as done in \cite{sun2023icomma,keetha2024splatam}. Although this is an intuitive way to solve $\boldsymbol{\rm T}_g$, we empirically find it leads to unsatisfactory performance in our case where the quality of $\boldsymbol{G}_t$ may degrade due to incremental accumulation without optimization.
Fortunately, we observe the pixels that are successfully matched are of high quality. In order to overcome the imperfect details in the rendering results, we propose to only focus on the matching pairs in 3D space for accurate camera pose optimization. 
The problem can be formulated as:
\begin{equation}
\hat{\mathbf{T}}=\underset{\mathbf{T} \in SE(3)}{\rm argmin}\ \ \mathcal{L}(\boldsymbol{\rm T} \ |\ \boldsymbol{I}_g, \boldsymbol{D}_{g}, \boldsymbol{\rm M}_g, \boldsymbol{G}_t)
\end{equation}
We iteratively optimize the pose $\boldsymbol{\rm T}$ to minimize the geometric discrepancy between rendering results and target image.
At each iteration of optimization, we leverage current $\boldsymbol{\rm T}$ for rendering and obtain the matched points in Euclidean space:
\begin{equation}
(\boldsymbol{x}_g, \boldsymbol{x}), (\boldsymbol{d}_g, \boldsymbol{d}) = \mathcal{M}(\boldsymbol{I}_g, \boldsymbol{D}_g, \mathcal{R}(\boldsymbol{G}_t\ |\ \boldsymbol{\rm M}_g, \boldsymbol{\rm T}))
\end{equation}
\begin{equation}
(\boldsymbol{X}_g, \boldsymbol{X}) = \text{Proj}^{-1}((\boldsymbol{x}_g, \boldsymbol{x}), (\boldsymbol{d}_g, \boldsymbol{d})\ |\ \boldsymbol{\rm M}_g)
\end{equation}
where $\mathcal{R}$ is differentiable rendering of color and depth. The matching and querying operation $\mathcal{M}$ first adopts LoFTR to get matching pair $(\boldsymbol{x}_g, \boldsymbol{x})$ between $\boldsymbol{I}_g$ and the rendered RGB image, and then queries the corresponding depth value $(\boldsymbol{d}_g, \boldsymbol{d})$ from $\boldsymbol{D}_g$ and the rendered depth respectively.
Then we formulate the optimization loss as:
\begin{equation}
\mathcal{L}=\frac{1}{Q} \sum_{i=0}^{Q-1}(|\boldsymbol{X}_g^i - \boldsymbol{X}^i|_2)
\end{equation}
where $Q$ is the number of matching pairs. Note that LoFTR predicts $(\boldsymbol{x}_g, \boldsymbol{x})$ in a differentiable way, so gradient can be backpropagated through both the rendered color and depth images.
In this way, we effectively align $\boldsymbol{\rm T}$ and $\boldsymbol{\rm T}_g$ by focusing on the most confident rendering results.

\begin{table*}[t]
\centering
\footnotesize
\caption{Image-goal Navigation Results. SR: Success Rate, SPL: Success weighted by Path Length. 
The best result in each column is \textbf{bold}, and the second best is \underline{underlined}.}
\vspace{-0.2cm}
\label{tab:image_goal}
\resizebox{0.97\textwidth}{!}{%
\begin{tabular}{@{}lcccccccccccccccc@{}}
\toprule
 & \multicolumn{8}{c}{\textbf{Straight}} & \multicolumn{8}{c}{\textbf{Curved}} \\
\cmidrule(lr){2-9}\cmidrule(lr){10-17}
\textbf{Method} 
& \multicolumn{2}{c}{\textbf{Easy}} 
& \multicolumn{2}{c}{\textbf{Medium}} 
& \multicolumn{2}{c}{\textbf{Hard}} 
& \multicolumn{2}{c}{\textbf{Overall}} 
& \multicolumn{2}{c}{\textbf{Easy}} 
& \multicolumn{2}{c}{\textbf{Medium}} 
& \multicolumn{2}{c}{\textbf{Hard}} 
& \multicolumn{2}{c}{\textbf{Overall}} \\
 & SR & SPL & SR & SPL & SR & SPL & SR & SPL 
 & SR & SPL & SR & SPL & SR & SPL & SR & SPL \\
\midrule
DDPPO~\cite{wijmans2019dd}                 
& 43.2 & 38.5 & 36.4 & 34.8 &  7.4 &  7.2 & 29.0 & 26.8
& 22.2 & 16.5 & 20.7 & 18.5 &  4.2 &  3.7 & 15.7 & 12.9 \\
NRNS~\cite{hahn2021no}                  
& 64.1 & 55.4 & 47.9 & 39.5 & 25.2 & 18.1 & 45.7 & 37.7
& 27.3 & 10.6 & 23.1 & 10.4 & 10.5 &  5.6 & 20.3 &  8.8 \\
ZSEL~\cite{al2022zero}               
& -    & -    & -    & -    & -    & -    & -    & -
& 41.0 & 28.2 & 27.3 & 13.9 & 18.6 &  9.3 & 25.9 & 17.6 \\
OVRL~\cite{yadav2023offline}               
& 53.6 & 34.7 & 48.6 & 33.3 & 32.5 & 21.9 & 44.9 & 30.0
& 53.6 & 31.8 & 47.6 & 30.2 & 35.6 & 22.0 & 45.6 & 28.0 \\
NRNS + SLING~\cite{wasserman2023last}       
& \underline{85.3} & 74.4 & 66.8 & 49.3 & 41.1 & 28.8 & 64.4 & 50.8
& 58.6 & 16.1 & 47.6 & 16.8 & 24.9 & 10.1 & 43.7 & 14.3 \\
OVRL + SLING~\cite{wasserman2023last}       
& 71.2 & 54.1 & 60.3 & 44.4 & 43.0 & 29.1 & 58.2 & 42.5
& 68.4 & 47.0 & 57.7 & 39.8 & 40.2 & 25.5 & 55.4 & 37.4 \\
RNR-Map~\cite{kwon2023renderable}
& 76.4 & 55.3 & \underline{73.6} & 46.1 & \underline{54.6} & 30.2 & \underline{68.2} & 43.9
& \underline{75.3} & \underline{52.5} & \underline{70.9} & \underline{42.3} & \underline{51.0} & \underline{27.4} & \underline{65.7} & \underline{40.8} \\
FeudalNav~\cite{johnson2024feudal}  
& 82.6 & \underline{75.0} & 71.0 & \underline{57.4} & 49.0 & \underline{34.2} & 67.5 & \underline{55.5}
& 72.5 & 51.3 & 64.4 & 40.7 & 43.7 & 25.3 & 60.2 & 39.1 \\
IGL-Nav (Ours)  
& \textbf{87.9} & \textbf{82.5} & \textbf{80.8} & \textbf{69.0} & \textbf{61.7} & \textbf{40.9} & \textbf{76.8} & \textbf{64.1}
& \textbf{82.8} & \textbf{77.7} & \textbf{80.7} & \textbf{70.0} & \textbf{57.0} & \textbf{39.6} & \textbf{73.5} & \textbf{62.4} \\
\bottomrule
\end{tabular}%
}
\vspace{-0.3cm}
\end{table*}

\subsection{Navigation}\label{sec:3.4}
We divide the navigation process into two stages: exploration based on coarse localization and target reaching based on fine localization. Figure \ref{fig:method3} illustrates the workflow of IGL-Nav. We will describe each stage in this section.

\textbf{Exploration for Target Discovery}.
When the agent is initialized in a new environment, its observations of the scene are insufficient. Therefore, we combine coarse target localization with frontier-based exploration to explore the scene and discover potential targets.
Based on the posed RGB-D inputs, we maintain an online occupancy map to indicate explored, unexplored and occupied area in BEV, where the frontiers of explored area can be computed.
At each time step, we select the nearest frontier to the agent and generate binary scores $S_f$ on the BEV map, where points on the selected frontier are set to 1, others are set to 0. We then project the activation map obtained in our coarse target localization module to BEV to get $S_a$.
We first filter the activation map $S_a$ by a threshold $\sigma_a$, setting scores below the threshold to zero. The agent then prioritizes exploring the location with the highest value in $S_a$. If all values in $S_a$ are zero, the agent selects the nearest location where the frontier score map $S_f$ equals 1.
We adopt Fast Marching Method~\cite{sethian1999fast} (FMM) for path planning and action generation given the to-be-explored location.

\textbf{Reaching Target}.
During exploration, the agent gradually approaches the target. We use the rendering-based stopper to determine if the target appears in agent's field of view. Once the target is detected, we switch to fine localization to compute the precise target pose. The XY coordinates of the computed pose is set to be destination, for which we apply FMM again for navigation.

\section{Experiment}   
    In this section, we first describe our experimental setting. Then we compare IGL-Nav with state-of-the-art image-goal navigation methods. Finally we conduct in-depth module-based analysis on our framework and further provide real-world deployment results.

\subsection{Experimental Setup}
We conduct experiments on image-goal navigation and the more challenging free-view image-goal navigation tasks.

\textbf{Datasets and Benchmarks}. For image-goal navigation, we follow the public Gibson~\cite{xia2018gibson} image-goal navigation dataset within the Habitat simulator~\cite{savva2019habitat} introduced by NRNS~\cite{hahn2021no}. The Gibson dataset includes 72 houses for training and 14 for validation. The NRNS dataset contains two path types (straight and curved), each with three difficulty levels (easy, medium, hard). For free-view image-goal navigation as introduced in Sec. \ref{sec:3.1}, we collect a large amount of data with Gibson for validation. Given the significant impact of the camera's field of view (FOV) on scene matching, we categorize our dataset into two FOV-based groups ($50^\circ \sim 75^\circ$ and $75^\circ \sim 100^\circ$), which can be intuitively understood as portrait and landscape orientations. Each category further includes three difficulty levels based on distance. Additionally, compared to the NRNS dataset, our free-view image-goal navigation dataset features target images captured from arbitrary angles and heights. Each of the six subsets contains 500 randomly sampled episodes.

\textbf{Compared Methods}. 
We compare IGL-Nav with existing state-of-the-art image-goal navigation methods~\cite{wijmans2019dd,hahn2021no,wasserman2023last,yadav2023offline,al2022zero,kwon2023renderable,johnson2024feudal}. For image-goal setting, we report results from the respective papers. For the proposed free-view image-goal setting, we evaluate open-sourced methods on this benchmark and compare with them.
Since some methods~\cite{wijmans2019dd,hahn2021no,wasserman2023last,yadav2023offline} only release test code, we perform zero-shot transfer to apply them to the new setting without retraining. We also report the zero-shot performance of IGL-Nav for fair comparison.
For methods~\cite{wijmans2019dd,hahn2021no} that provide training scripts, we train them on the free-view image-goal navigation data for comparison.

\subsection{Comparison with State-of-the-art}
We compare with state-of-the-art image-goal navigation methods on the two benchmarks described above.
Table \ref{tab:image_goal} demonstrates the results on image-goal navigation task. IGL-Nav establishes new state-of-the-art performance and outperforms previous methods by a large margin on all metrics, which validates the effectiveness of 3D gaussian representation and the proposed coarse-to-fine target localization strategy for image-goal navigation.

\begin{table*}[t]
\centering
\footnotesize
\caption{Free-view Image-goal Navigation Results. SR: Success Rate, SPL: Success weighted by Path Length.}
\vspace{-0.2cm}
\label{tab:image_goal2}
\resizebox{0.97\textwidth}{!}{%
\begin{tabular}{@{}lcccccccccccccccc@{}}
\toprule
 & \multicolumn{8}{c}{\textbf{Narrow FOV ($50^\circ \sim 75^\circ$)}} & \multicolumn{8}{c}{\textbf{Wide FOV ($75^\circ \sim 100^\circ$)}} \\
\cmidrule(lr){2-9}\cmidrule(lr){10-17}
\textbf{Method} 
& \multicolumn{2}{c}{\textbf{Easy}} 
& \multicolumn{2}{c}{\textbf{Medium}} 
& \multicolumn{2}{c}{\textbf{Hard}} 
& \multicolumn{2}{c}{\textbf{Overall}} 
& \multicolumn{2}{c}{\textbf{Easy}} 
& \multicolumn{2}{c}{\textbf{Medium}} 
& \multicolumn{2}{c}{\textbf{Hard}} 
& \multicolumn{2}{c}{\textbf{Overall}} \\
 & SR & SPL & SR & SPL & SR & SPL & SR & SPL 
 & SR & SPL & SR & SPL & SR & SPL & SR & SPL \\
\midrule
\multicolumn{8}{l}{\textbf{\textit{Zero-shot Transfer (Training on Image-goal Navigation Data)} }} \\
\midrule
DDPPO~\cite{wijmans2019dd}               
& 15.8 & 10.5 & 9.6 & 7.2 & 5.4 & 3.1 & 10.3 & 6.9 
& 20.2 & 16.5 & 16.6 & 12.5 & 9.8 & 5.7 & 15.5 & 11.6 \\
NRNS~\cite{hahn2021no}               
& 19.8 & 10.6 & 15.8 & 9.0 & 7.8 & 4.0 & 14.5 & 7.9
& 28.4 & 16.6 & 21.2 & 14.5 & 10.6 & 5.9 & 20.1 & 12.3 \\
OVRL~\cite{yadav2023offline}              
& 23.8 & 16.6 & 19.2 & 10.5 & 8.2 & 6.9 & 17.1 & 11.3
& 27.6 & 19.2 & 22.8 & 12.6 & 14.8 & \underline{8.6} & 21.7 & 13.5 \\
NRNS + SLING~\cite{wasserman2023last}       
& \underline{32.8} & 15.3 & \underline{23.6} & 13.2 & 9.8 & 5.6 & \underline{22.1} & 11.4
& \underline{38.6} & 19.1 & \underline{32.6} & \underline{18.5} & \underline{17.2} & 8.3 & \underline{29.5} & 15.3 \\
OVRL + SLING~\cite{wasserman2023last}          
& 28.2 & \underline{20.1} & 23.2 & \underline{18.7} & \underline{11.8} & \underline{7.1} & 21.1 & \underline{15.3}
& 36.4 & \underline{25.9} & 31.6 & \underline{18.5} & 15.2 & 7.6 & 27.7 & \underline{17.3} \\
IGL-Nav (Ours)  
& \textbf{53.2} & \textbf{45.1} & \textbf{47.8} & \textbf{40.5} & \textbf{28.2} & \textbf{22.0} & \textbf{43.1} & \textbf{35.9}
& \textbf{56.2} & \textbf{48.3} & \textbf{55.2} & \textbf{46.1} & \textbf{30.8} & \textbf{23.9} & \textbf{47.4} & \textbf{39.4} \\
\midrule
\multicolumn{8}{l}{\textbf{\textit{Supervised (Training on Free-view Image-goal Navigation Data)} }} \\
\midrule
BC + GRU              
& 13.2 & 6.8 & 10.4 & 8.8 & 6.6 & 4.9 & 10.1 & 6.8
& 22.0 & 14.4 & 16.0 & 11.4 & 9.2 & 6.9 & 15.7 & 10.9 \\
BC + Metric Map              
& 22.8 & 15.9 & 20.6 & 15.6 & 7.4 & 5.2 & 16.9 & 12.2
& 25.4 & 19.5 & 22.8 & 18.5 & 4.8 & 3.5 & 17.7 & 13.8 \\
DDPPO~\cite{wijmans2019dd}           
& 19.4 & 11.3 & 16.4 & 10.4 & 9.6 & 6.0 & 15.1 & 9.2
& 26.8 & 17.8 & 19.0 & 12.4 & 15.6 & 9.8 & 20.5 & 13.3 \\
NRNS~\cite{hahn2021no}                
& 30.8 & 24.4 & 27.8 & \underline{24.5} & 11.2 & 8.9 & 23.3 & 19.3
& 39.6 & \underline{35.0} & 35.8 & 30.1 & 13.8 & 8.3 & 29.7 & 24.5 \\
NRNS + SLING~\cite{wasserman2023last}           
& \underline{40.2} & \underline{31.8} & \underline{37.2} & 23.9 & \underline{19.8} & \underline{9.8} & \underline{32.4} & \underline{21.8}
& \underline{49.8} & \underline{35.0} & \underline{40.8} & \underline{30.4} & \underline{21.6} & \underline{12.7} & \underline{37.4} & \underline{26.0} \\
IGL-Nav (Ours)  
& \textbf{70.4} & \textbf{64.2} & \textbf{60.6} & \textbf{51.4} & \textbf{40.0} & \textbf{28.9} & \textbf{57.0} & \textbf{48.2}
& \textbf{77.2} & \textbf{73.1} & \textbf{69.8} & \textbf{60.1} & \textbf{42.8} & \textbf{31.9} & \textbf{63.3} & \textbf{55.0} \\
\bottomrule
\end{tabular}%
}
\vspace{-0.25cm}
\end{table*}

The results on free-view image-goal navigation task is shown in Table \ref{tab:image_goal2}. As this task is much more challenging than conventional image-goal setting, we observe a significant performance drop on each metric. When directly transferred from image-goal to free-view image-goal setting, IGL-Nav still maintains a huge performance lead compared with other state-of-the-art methods. The performance of IGL-Nav can be further boosted with training data on the free-view image-goal task.
Note that the zero-shot transferring performance of IGL-Nav is even better than other methods under supervised setting, which demonstrates the great generalization ability of our approach.

\begin{table}[t]
\centering
\footnotesize
\caption{Performance of IGL-Nav when depth and camera intrinsics are unavailable.}
\vspace{-0.25cm}
\label{tab:pred_depth}
\resizebox{0.45\textwidth}{!}{%
\begin{tabular}{ccccc}
\toprule
\multirow{2}{*}{\textbf{Method}} & \multicolumn{2}{c}{\textbf{Narrow FOV ($50^\circ \sim 75^\circ$)}} & \multicolumn{2}{c}{\textbf{Wide FOV ($75^\circ \sim 100^\circ$)}} \\
~ & \hspace{6mm}SR & SPL & \hspace{6mm}SR & SPL \\
\midrule
Predicted Depth & \hspace{6mm}53.8 & 44.7 & \hspace{6mm}61.0 & 51.7 \\
Measured Depth & \hspace{6mm}\textbf{57.0} & \textbf{48.2} & \hspace{6mm}\textbf{63.3} & \textbf{55.0} \\
\bottomrule
\end{tabular}%
}
\vspace{-0.15cm}
\end{table}

\begin{figure}
% \vspace{-.3cm}
  \centering
  \includegraphics[width=0.477\textwidth]{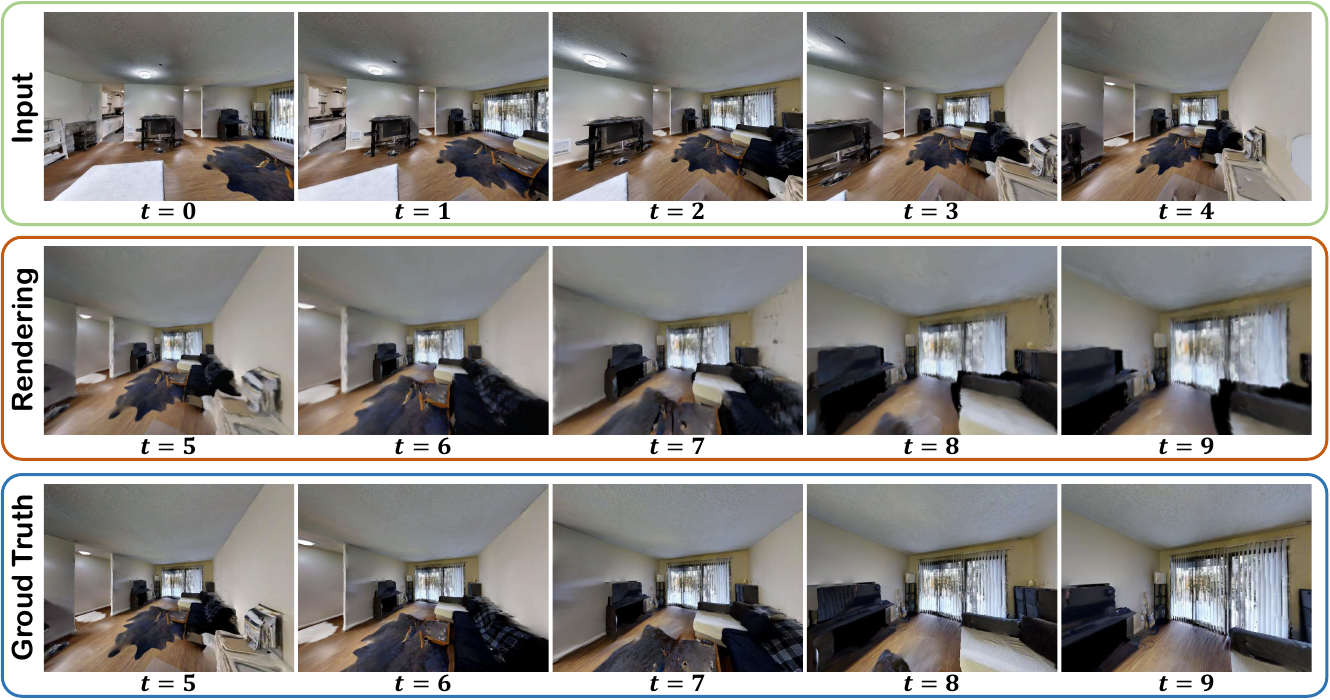}
  \vspace{-.65cm}
  \caption{Rendering results of our incremental 3DGS.}
  \label{fig:vis_render}
  \vspace{-.45cm}
\end{figure}

\subsection{Analysis of IGL-Nav}
We further conduct in-depth module-by-module analysis on our IGL-Nav framework with sufficient visualization results and ablation studies, which is divided into three parts according to our module design. All ablation studies are conducted on the free-view image-goal setting.

\textbf{Incremental 3DGS Prediction}.
Following the setting of RNR-Map~\cite{kwon2023renderable}, we assume depth information and camera intrinsics are known in our experiments. When these information is unavailable, we can simply adopt a depth estimator~\cite{piccinelli2024unidepth} to predict them. As shown in Table \ref{tab:pred_depth}, with predicted depth and camera intrinsics, the performance of IGL-Nav is still robust.
We further visualize rendering results of our 3DGS representation in Figure \ref{fig:vis_render}. Although maintained in an incremental and feed-forward manner, our 3DGS still demonstrates photorealistic novel view synthesis capability.

\textbf{Coarse Target Localization}.
In our coarse localization module, the sphere space is discretized with a regular icosahedron and its $\gamma$-level fractal. A larger value of $\gamma$ leads to finer discretization of the spherical surface and results in a greater number of convolution kernels $N$. We study the the effects of different levels to the final performance in Table \ref{tab:level}. It is shown that using a 3-level subdivision achieves best performance, because a finer discretization will reduce quantization error and improve the accuracy of coarse localization. However, a larger $\gamma$ results in high computational cost, making training inefficient.

\begin{table}[t]
\centering
\footnotesize
\caption{Effects of different subdivision levels in coarse target localization to the final performance.}
\vspace{-0.25cm}
\label{tab:level}
\resizebox{0.45\textwidth}{!}{%
\begin{tabular}{ccccc}
\toprule
\multirow{2}{*}{\textbf{Level} ($\gamma$)} & \multicolumn{2}{c}{\textbf{Narrow FOV ($50^\circ \sim 75^\circ$)}} & \multicolumn{2}{c}{\textbf{Wide FOV ($75^\circ \sim 100^\circ$)}} \\
~ & \hspace{6mm}SR & SPL & \hspace{6mm}SR & SPL \\
\midrule
1 & \hspace{6mm}19.7 & 12.0 & \hspace{6mm}24.9 & 16.8 \\
2 & \hspace{6mm}41.3 & 34.4 & \hspace{6mm}48.9 & 42.1 \\
3 & \hspace{6mm}\textbf{57.0} & \textbf{48.2} & \hspace{6mm}\textbf{63.3} & \textbf{55.0} \\
\bottomrule
\end{tabular}%
}
\vspace{-0.1cm}
\end{table}

\begin{table}[t]
\centering
\footnotesize
\caption{Effects of different stoppers in fine target localization to the final performance.}
\vspace{-0.25cm}
\label{tab:stopper}
\resizebox{0.45\textwidth}{!}{%
\begin{tabular}{ccccc}
\toprule
\multirow{2}{*}{\textbf{Stopper}} & \multicolumn{2}{c}{\textbf{Narrow FOV ($50^\circ \sim 75^\circ$)}} & \multicolumn{2}{c}{\textbf{Wide FOV ($75^\circ \sim 100^\circ$)}} \\
~ & \hspace{6mm}SR & SPL & \hspace{6mm}SR & SPL \\
\midrule
IGL-Nav w/out Stopper & \hspace{6mm}45.7 & 32.9 & \hspace{6mm}46.2 & 37.6 \\
IGL-Nav w/ SLING~\cite{wasserman2023last}    & \hspace{6mm}49.0 & 40.7 & \hspace{6mm}52.4 & 45.0 \\
IGL-Nav & \hspace{6mm}\textbf{57.0} & \textbf{48.2} & \hspace{6mm}\textbf{63.3} & \textbf{55.0} \\
\bottomrule
\end{tabular}%
}
\vspace{-0.3cm}
\end{table}

\begin{figure*}
% \vspace{-.3cm}
  \centering
  \includegraphics[width=0.96\textwidth]{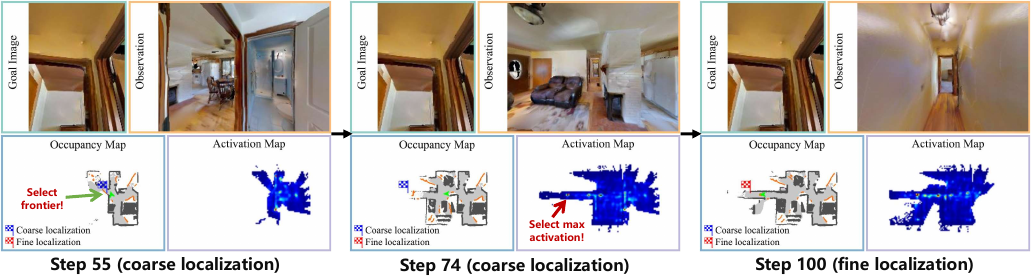}
  \vspace{-.3cm}
  \caption{Visualization of navigation process in Habitat simulator.}
  \label{fig:vis_nav_sim}
  \vspace{-.2cm}
\end{figure*}

\begin{figure*}
% \vspace{-.3cm}
  \centering
  \includegraphics[width=0.96\textwidth]{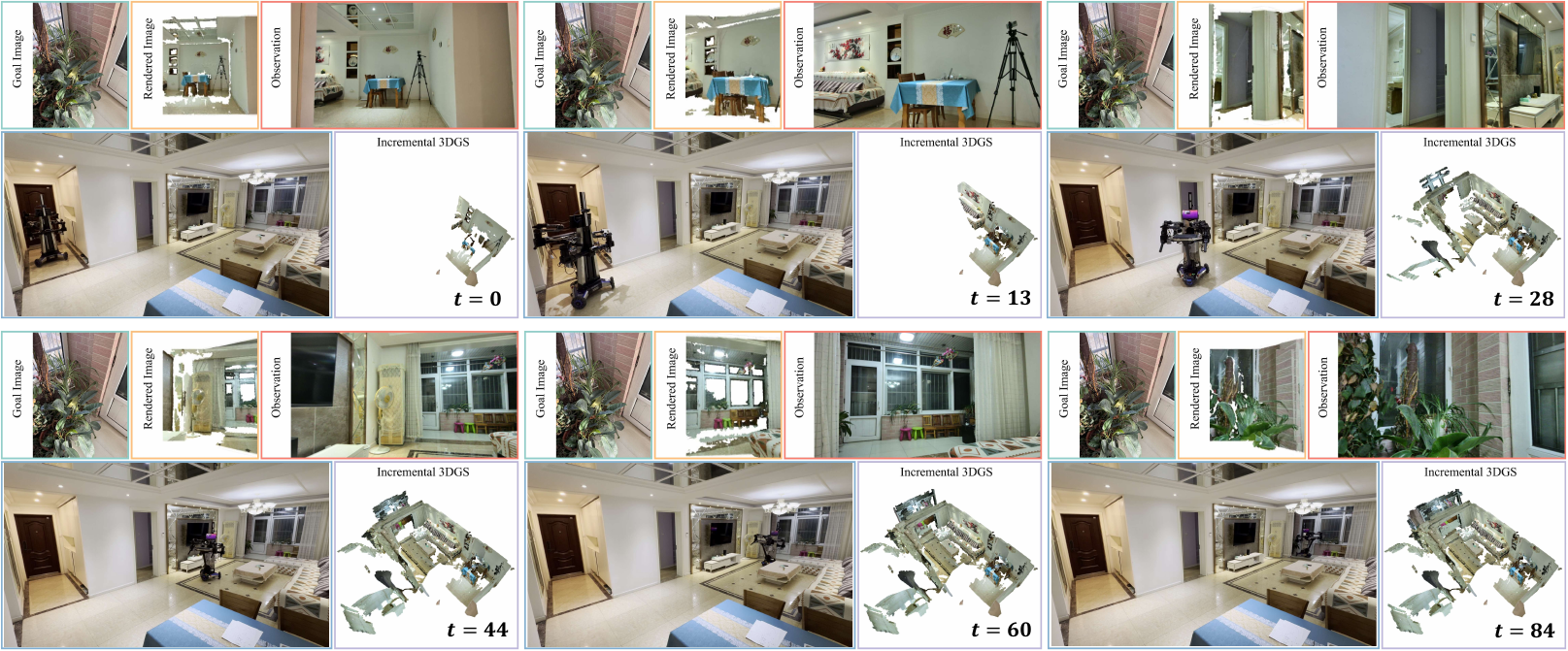}
  \vspace{-.2cm}
  \caption{Visualization of navigation process in the real world. The agent is successfully guided to a free-view goal image captured by a cellphone in complex indoor environments. IGL-Nav exhibits strong generalization ability and sim-to-real transfer performance.
}
  \label{fig:real_world}
  \vspace{-.2cm}
\end{figure*}

\textbf{Fine Target Localization}.
We compare different stoppers in Table \ref{tab:stopper}. The first row refers to only using coarse target localization, and the second row refers to using the widely adopted SLING~\cite{wasserman2023last} as the stopper and fine localization module. It is shown that our 3DGS-based stopper and matching-constrained optimization is more suitable for the overall navigation system of IGL-Nav.

We also visualize the navigation process in Figure \ref{fig:vis_nav_sim}. The agent is guided with frontier location, activation map obtained with 3D convolution and iterative pose optimization during the exploration. It is shown that our IGL-Nav can effectively localize the target image even with partial observation, and accurately guide the agent to final location with fine-grained rendering-based optimization.

\subsection{Real-world Deployment}
We further deploy IGL-Nav on real-world robotic platform to test its generalization ability. The model is directly taken from the free-view image-goal setting (supervised) without any finetuning on real-world data. As shown in Figure \ref{fig:teaser} and \ref{fig:real_world}, we use a cellphone to capture the target image in a viewpoint that is unreachable by the robotic agent's camera. Benefit from the flexible rendering capability of 3DGS representation, the agent effectively reaches this free-view goal with the coarse-to-fine localization method.

\section{Conclusion}   
    In this paper , we have proposed IGL-Nav for efficient and 3D-aware image-goal navigation. We incrementally maintain a 3DGS scene representation in feed-forward manner, which is then utilized for coarse-to-fine target localization. We analyze the pose space of the goal image and discretize both the pose space and scene embedding to apply efficient 3D convolution-based coarse matching. When the agent is close to the goal, we switch to fine localization by optimizing the camera pose via differentiable rendering on the confident matching pairs.
The proposed IGL-Nav significantly outperforms existing state-of-the-art methods on image-goal and free-view image-goal settings. Real-world experiments further demonstrate our generalization ability.
A limitation of IGL-Nav is that it requires depth and camera intrinsics of goal image. However, as we show in experiments, using existing monocular depth estimation~\cite{piccinelli2024unidepth} to predict them can satisfactorily solve this problem.

% \section*{Acknowledgement}
% This work was supported in part by the National Natural Science Foundation of China under Grant 624B2076, 62376132, and 62321005.

{
    \small
    \bibliographystyle{ieeenat_fullname}
    \bibliography{main}
}

\end{document}